\documentclass[11pt]{article}
\usepackage{nodalida2021}
\usepackage{times}
\usepackage{url}
\usepackage{latexsym}
\usepackage[utf8]{inputenc}
\usepackage{graphicx}
\usepackage{multirow}
\usepackage{appendix}
\usepackage{float}
\usepackage[table,xcdraw]{xcolor}

\csname @openrightfalse\endcsname

\aclfinalcopy 

\title{SELF \& FEIL: Emotion and Intensity Lexicons for Finnish}
\author{Emily Öhman \\
University of Helsinki \\
{\tt emily.ohman@helsinki.fi}}

\date{September 2020}

\begin{document}

\maketitle
\begin{abstract}
    This paper introduces a Sentiment and Emotion Lexicon for Finnish (SELF) and a Finnish Emotion Intensity Lexicon (FEIL). We describe the lexicon creation process and evaluate the lexicon using some commonly available tools. The lexicon uses annotations projected from the NRC Emotion Lexicon with carefully edited translations. To our knowledge, this is the first comprehensive sentiment and emotion lexicon for Finnish.
\end{abstract}
\section{Introduction}
Emotion and sentiment lexicons are used in sentiment analysis and emotion detection and their downstream applications. They can be used ``as is" in purely lexicon-based approaches or as a feature extraction tool for data-driven classifiers. 

Many resources for sentiment analysis and emotion detection exist, but most of them are for English. The NRC Emotion Lexicon \cite{mohammad2010emotions, Mohammad13} is one of the few exceptions. Although it was originally annotated for English only, the English words have been translated using Google Translate and there are now 14,182 entries in total translated into 104 languages.

The NRC Emotion Lexicon is used in many applications such as fake news detection \cite{vosoughi2018spread} and in detecting trajectories of plot movement \cite{jockers2015syuzhet}. The evaluation of a lexicon is not straightforward and can mainly be determined in terms of usefulness, specifically in downstream tasks. Therefore our evaluation focus is on showing how SELF and FEIL improve the results for Finnish language tasks as compared to the original machine translated NRC Lexicons.

The SELF and FEIL lexicons are available on GitHub\footnote{\url{https://github.com/Helsinki-NLP/SELF-FEIL}}
, with plans to both update the existing ones based on user feedback, as well as create new ones with domain-specific alterations.

\section{Background}
Emotion lexicons can be created automatically or manually, but both approaches often, at least partially, utilize dictionaries that list emotion categories in some way. For automatic creation no manual verification is applied. Instead, other features are used to determine the sentiment or emotion. For example, \newcite{Mayu2017auto} automatically created an emoji-lexicon by leveraging cosine similarity values and co-occurence from WordNet Affect \cite{wordnet}. For manual creation human annotators are used by asking them to determine how specific words relate to different sentiments and/or emotions. In this paper we focus on the manual creation of lexicons, specifically the augmentation of existing ones that were originally created manually and later automatically translated.

The first version of the NRC Emotion Lexicon (a.k.a. EmoLex) \cite{mohammad2010emotions} was created using Mechanical Turk for crowd-sourcing annotations. The lexicon was later significantly augmented to its current 14,182 lexical items also using Mechanical Turk \cite{Mohammad13}. For the first version, the annotators were asked to annotate for emotions \textbf{evoked} by the words, and for the later version, they were asked to annotate for emotions \textbf{associated} with the words as the developers of the lexicon discovered this led to better agreement scores.

Some of the multilingual support of the NRC Emotion Lexicon has been tested previously for at least Finnish, Spanish, Portuguese, and Arabic \cite{ohman2016challenges, salameh2015sentiment,MohammadSK16} to evaluate emotion preservation in translation.

The NRC Emotion Intensity Lexicon \cite{LREC18-AIL} to a large extent uses the same data as the emotion lexicon, but the annotations were compiled using best-worst scaling (BWS). BWS is an efficient method to collect massive amounts of scaled annotations and has been proven to beat rating scales and other methods in both quality and cost\footnote{Cost because other methods take exponentially more time to collect the same amount of annotations.} \cite{kiritchenko-mohammad-2017-best}. The main difference between the intensity lexicon and the emotion lexicon is that the intensity lexicon does not include sentiments (positive, negative), and that instead of a Boolean option for each emotion (0 for not associated with a particular emotion, 1 for being associated with a particular emotion) the intensity lexicon gives a score as to how intense the emotion associated with a word is (a score between 0 and 1, with 0 for no association and therefore no intensity, to 1 for the highest intensity).

There are many considerations that need to be taken into account when using resources created for one language with another language. Emotion words are closely linked with the culture they are spoken in and emotions and feelings are expressed quite differently in different cultures and languages. Although research shows the universality of affect categories \cite{Cowen2019ThePO, Scherer1994EvidenceFU}, \newcite{MohammadSK16} list several error types when translating emotion words. These are, e.g., mistranslation, cultural differences and different word connotations, as well as different sense distributions. When projecting annotations, particularly emotion annotations, from one language to another, it is important to consider all of these aspects.

\textit{Syuzhet} \cite{jockers2017cran}, an R package used to create graphs approximating plot movement in novels using emotion dictionaries and other tools, is used in this paper to evaluate the impact of the changes made to the lexicon as compared to the Google Translate version. It has built-in options for some common emotion lexicons and supports custom lexicons as well. It is a useful tool for comparing the impact of the changes to the lexicon in a practical approach that can be verified by human experts.

\section{Lexicon Creation and Description}

The automatic Finnish translations of the NRC Lexicons were re-translated with the most current version of the Google Translate plugin for Google Sheets. First, within these translations, duplicates were marked and all translations were carefully evaluated to match the English word's meaning and the associated emotions, and in the case of the FEIL lexicon, intensity. The problem with Google Translate is that it chooses the most common translation, especially when translating individual words rather than words in context, which means that with an emotion dictionary with many synonyms and near-synonyms, Google Translate provides the same common word for all as the default translation.

Naturally, this leads to issues with representativity as the Finnish lexicon, if left as is, would only represent the most common words in Finnish, disregarding any synonyms or near-synonyms leading to a loss of nuances and insufficient coverage of emotion words. Hence, those duplicates were carefully examined in order to find alternative translations that matched the original word better in terms of meaning and connotation, both by human experts and synonym dictionaries and thesauruses (see table \ref{tab:hurskas} for an example). The original lexicon also included several alternative spelling options, which were translated as the same word. Such duplicates  (e.g. tumor/tumour) were also removed.

Another type of over-generalization was also discovered: connotative-generalization (the meaning of the original word was over-generalized so that the translation had lost all original connotations (see the example of `emaciated' in table \ref{tab:corrections-overview} in the Appendix).

\begin{table}[h]
\centering
\resizebox{\columnwidth}{!}{
\begin{tabular}{lll}
\rowcolor[HTML]{EFEFEF} 
Original English & Google Translate & Edited Translation \\
pious            & hurskas          & hurskas            \\
devout           & hurskas          & harras             \\
saintly          & hurskas          & pyhimysmäinen      \\
godly            & hurskas          & jumalinen        
\end{tabular}}
\caption{Example of lexicon editing: the case of \textit{hurskas}}
\label{tab:hurskas}
\end{table}

The reverse was also found with some instances where English has a more fine-grained separation of some concepts where Finnish does not. E.g. Finnish does not differentiate between \textit{poison-venom-toxin}. It is all the same word \textit{myrkky}. Some English words can be both nouns and verbs, and therefore, if not specified it is impossible to tell which one is being evaluated\footnote{In the original task for English, the annotators were given a test for each word to check that they understood the meaning of the word as intended. This is information is not available in the published dictionary.} or if both are. One such word was \textit{rape}. In this case the verb form was added with the same emotion and intensity associations as the noun form. There were also cases of clear mistranslations, some of which occurred with polysemic and homonymous words where the Finnish translation was clearly not the one meant based on the associated emotions and/or intensities. For example \textit{birch} had been translated as `koivu', a birch tree, but the negative associations suggested that what had been meant was the act of flogging. In many cases, the easiest solution was to remove ambiguous entries. See table \ref{tab:corrections-overview} in the Appendix for an overview of common corrections.

For \textit{emaciated}, the automatic translation was too general. The corrected form is less common but almost identical in meaning and connotation to the original English. As the term \textit{rabble} has such negative connotations in English, the automatic translation was much too neutral and was changed to a word with similar connotations. As for \textit{corroborate} the issue was with there not existing a one-word translation for \textit{corroborate} in Finnish and the meaning of the automatically translated word being much closer to \textit{strengthen}. The meaning of the original English words \textit{cede, relinquish} are somewhat synonymous so it makes sense to have both words in the English lexicon with nearly identical emotional intensities, but there is only one one-word translation for Finnish so the duplicate was removed. Another example is \textit{furor} which in English had anger intensities of 0.9 and in Finnish had been translated into \textit{villitys - fad}, but the best translation was already paired with \textit{rage} and as no suitable alternative translation was found, the entry was deleted.

The best translation is not always necessarily the best choice for a lexicon entry. If one translated word truly is the best translation for several English words, an evaluation of the best match needs to be made. If possible, the duplicate is not removed, but altered to still match the English word, but using a different word, even if not as accurately translated.

Cultural differences were also evident in the lexicon. North-Americans are on average more religious than the Nordic people \cite{skirbekk2015future}, which means that many religious words seem to have much more positive connotations in the US and Canada than they do in Northern Europe, and Scandinavia in particular (see the example of \textit{hurskas} in table \ref{tab:hurskas}). This was most evident with religious words, but other cultural connotations seemed to be present as well. In these cases though, unless the association was glaringly wrong, they were kept as is: It is a slippery slope to try and push one's own judgment onto the intensity scores and emotion-word associations. 
Everyone is inherently biased and relies on their own experience when making judgments. Only when enough people agree on a judgment, can it be seen as culturally representative. Since in this case only one person did most of the corrections, words were rather deleted than letting a lone annotator's subjective judgment overly influence the lexicon.

The final distribution of the SELF and FEIL lexicons can be seen in table \ref{tab:SELF} in the Appendix. All the adjustments mentioned in this section resulted in an overall reduction in lexicon size by 12.2\% from 14182 word-emotion association pairs to 12448 entries in Finnish. The final distribution of the FEIL lexicon with all the adjustments mentioned in this section resulted in an overall reduction in lexicon size by 10.5\% from 8149 intensities to 7291 entries for Finnish.

The distribution of the types of corrections is presented in table \ref{tab:errors}. It is easier to detect sense and specificity  mistranslations in the intensity lexicon when the intensities of the associated emotions are clear. This is likely the reason for the higher relative occurrence of such corrections in the intensity lexicon. However, these types of errors are also the hardest to detect overall, and therefore, this is the category that is most likely to increase the most when the lexicon is updated and revised further.
\begin{table*}[htbp!]
\centering
\begin{tabular}{llll}
\rowcolor[HTML]{EFEFEF} 
\% of corrections & SELF & FEIL & Ex.                                                    \\
Duplicate removal           & 46.1\%                  & 41.3\% & identical target words with no alternative translation \\
Duplicate replacement       & 36.8\%                  & 33.4\% & identical target words with alternative translation    \\
Mistranslation -sense       & 2.0\%                   & 4.9\%  & \textit{birch} to birch tree instead of flogging                \\
Mistranslation -specificity & 3.7\%                   & 6.3\%  & \textit{emaciated} to \textit{laihtunut} instead of \textit{riutunut}             \\
Grammatical difference      & 1.5\%                   & 1.2\%  & part-of-speech difference                              \\
Cultural difference         & 0.9\%                   & 1.7\%  & overly positive connotations of religious words        \\
Other, undefined          & 9.0\%                     & 11.2\% &                                                       
\end{tabular}
\caption{Percentage of different types of corrections in SELF and FEIL}
\label{tab:errors}
\end{table*}

\section{Lexicon Evaluation}
The lexicon was evaluated by having an expert in Finnish literature evaluate the affect and mood of the novel Rautatie by Juhani Aho.  

\begin{figure*}[htbp!]
\centering
\includegraphics[scale=0.35]{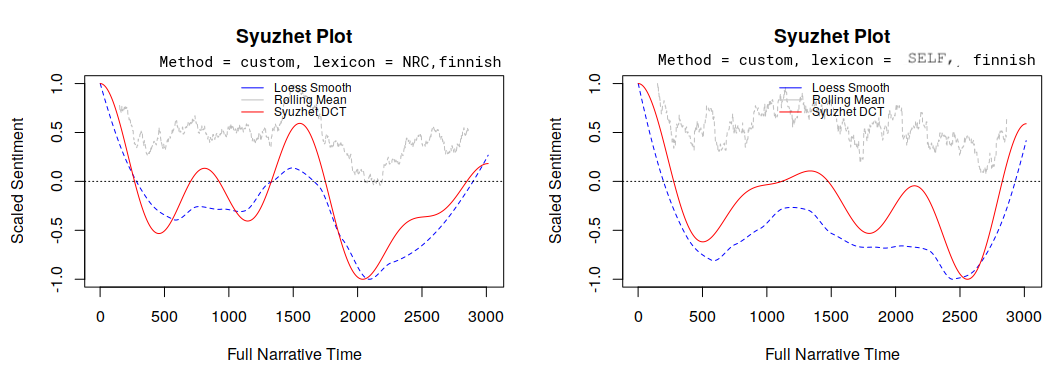}
\caption{\textit{syuzhet} results for Rautatie using different lexicons}
\label{rautatie_all}
\end{figure*}

This qualitative evaluation was then compared with both the original version of the NRC Emotion Lexicon and SELF using \textit{syuzhet} on \textit{Rautatie}, downloaded from Project Gutenberg. The results can be seen in figure \ref{rautatie_all} where on the left is the NRC Emotion Lexicon and on the right SELF. The two plots show similar overall patterns, but are decidedly different in detail, with SELF providing the more accurate plot when evaluated by the expert.

The emotion word distributions (see table \ref{tab:lexcomp} in Appendix) reveal that in general the SELF lexicon finds more emotion words in the novel, but that \textit{anger, disgust}, and \textit{sadness}, and therefore also \textit{negative} were over-represented in the original lexicon, likely by duplicates. Another surprising find is the high hit rates for \textit{trust}. This is probably due to the fact that the most common \textit{trust} word in the novel is \textit{rautatie - railroad}, a.k.a. the title and main theme of the novel. 

\begin{figure}[htbp!]
\centering
\includegraphics[scale=0.31]{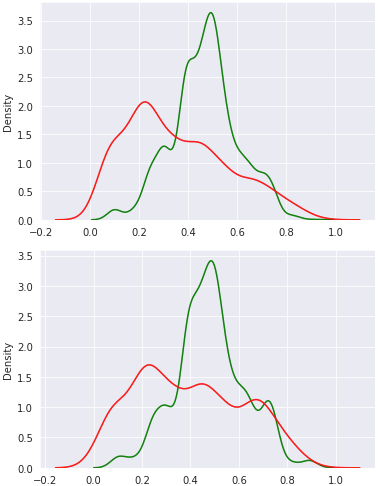}
\caption{Intensity results (KDE) for \textit{Rautatie} using FEIL (top) and NRC (bottom)}
\label{FEILeval}
\end{figure}

The intensity word distribution in the NRC Intensity Lexicon and FEIL can be compared in figure \ref{FEILeval} where the NRC Emotion Intensity Lexicon and FEIL have been matched with \textit{Rautatie} presented by Kernel Density Estimates (KDE). The differences are more subtle than with SELF (table \ref{tab:lexcomp}). The KDE curves show similar peaks for both lexicons, but with many more matches with FEIL (top) than the NRC lexicon (bottom). The main difference is the more linear drop towards high intensity negative items in FEIL compared to a plateau between 0.5 and 0.8 for negative items in the NRC lexicon evaluation of \textit{Rautatie}. The literary expert agreed that the plot created using SELF resembled the actual plot development of the novel more closely.

\section{Concluding Discussion}
\begin{samepage}
The SELF and FEIL lexicons have proved to be valuable augmentations of the NRC Emotion and Intensity Lexicons. The problems with the use of the NRC lexicons' automatically translated versions do not seem to be caused by any issues with the original annotations (also attested by the thousands of projects that have successfully used them for English), nor with the annotation projection as such, but mostly by Google Translate's algorithms which are not optimal for translating single words out of context. 
\nopagebreak
SELF and FEIL will hopefully prove to be useful in advancing Finnish emotion detection and sentiment analysis research. 
\end{samepage}

\section*{Acknowledgements}
I would like to thank Dr. Saif Mohammad for taking the time to discuss the SELF and FEIL lexicons with me, and for his helpful suggestions in improving the draft version of this paper.
\bibliographystyle{acl_natbib}
\bibliography{diss,arts,SELF}
\begingroup
\let\clearpage\relax
\appendix

\section{Appendix}

\begin{table*}[]
\centering
\begin{tabular}{lll}
\rowcolor[HTML]{C0C0C0} 
Original English                               & Automatic Translation                               & Corrected Form(s) \\
birch                                          & koivu                                               & -entry removed-   \\
\rowcolor[HTML]{EFEFEF} 
emaciated                                      & laihtunut (having lost weight)                      & riutunut          \\
rabble                                         & lauma (herd, flock)                                 & roskaväki         \\
\rowcolor[HTML]{EFEFEF} 
corroborate                                    & \cellcolor[HTML]{EFEFEF}                            & -entry removed-   \\
\rowcolor[HTML]{EFEFEF} 
strengthen                                     & \multirow{-2}{*}{\cellcolor[HTML]{EFEFEF}vahvistaa} & -entry kept-      \\
cede                                           &                                                     & -entry kept-      \\
relinquish                                     & \multirow{-2}{*}{luovuttaa (to give up)}            & -entry removed-   \\
\rowcolor[HTML]{EFEFEF} 
\cellcolor[HTML]{EFEFEF}                       & raiskaus (N)                                        & -entry kept-      \\
\rowcolor[HTML]{EFEFEF} 
\multirow{-2}{*}{\cellcolor[HTML]{EFEFEF}rape} & raiskata (V)                                        & -entry added-    
\end{tabular}
\caption{Examples of fixes to the SELF and FEIL lexicons}
\label{tab:corrections-overview}
\end{table*}

\begin{table*}[]
\centering
\resizebox{\textwidth}{!}
{\begin{tabular}{ccccccccccc}
\rowcolor[HTML]{EFEFEF} 
positive & negative & anger & anticipation & disgust & fear & joy & sadness & surprise & trust & Lexicon \\
2117     & 2938     & 1084  & 783          & 919     & 1309 & 636 & 1059    & 473      & 1130 & SELF \\
&&1304  & 805          & 946     & 1554 & 1145 & 183 &    & 1354 & FEIL
\end{tabular}}
\caption{Distribution of Emotions in SELF \& FEIL after corrections.}
\label{tab:SELF}
\end{table*}

\begin{table*}[]
\resizebox{\textwidth}{!}{

\begin{tabular}{ll|llllllll|l}
\rowcolor[HTML]{EFEFEF} 
positive & negative & anger & anticipation & disgust & fear & joy & sadness & surprise & trust &      \\ \hline
1153     & 1258     & 299   & 482          & 303     & 506  & 345 & 368     & 248      & 584   & SELF \\
828      & 1363     & 481   & 299          & 381     & 433  & 241 & 451     & 162      & 321   & NRC 
\end{tabular}}
\caption{Comparison of lexicon matches for \textit{Rautatie} by Juhani Aho}
\label{tab:lexcomp}

\end{table*}
\endgroup

\end{document}